\documentclass{article}
\usepackage{spconf,amsmath,graphicx,hyperref,amsfonts}
\usepackage{authblk}
\usepackage{xcolor}
\usepackage{tikz}
\usetikzlibrary{shapes.geometric,arrows.meta,positioning,fit,backgrounds,calc}
\usepackage[scaled=0.92]{helvet}
\usepackage{cleveref}
\usepackage{adjustbox}
\usepackage{enumitem}

\makeatletter
\renewcommand{\name}[1]{\gdef\@name{{\em #1}}}
\makeatother

\title{Data-Driven Variable-Length Compression of 3D Representations}
\name{
Yvette Y. Lin $^{*}$ \qquad
Sai Karthikey Pentapati $^{*2}$ \qquad
Sarvesh Sundaram$^{1}$ \\[-0.1em]
Guandao Yang$^{2}$ \qquad
Sara Fridovich-Keil$^{1}$
}

\address{\vspace{0.7em}
$^{1}$ Georgia Institute of Technology \qquad
$^{2}$ The University of Texas at Austin
}
\begin{document}

\maketitle

{\let\thefootnote\relax\footnotetext{$^{*}$Equal contribution.}}

\begin{abstract}
Compressing 3D shapes is important for the efficiency of many applications, yet existing methods either cannot leverage dataset-level statistics or use a fixed compression ratio for all shapes. We propose a learned variable-length compression method that exploits structure across a training dataset while adapting compressed representation size to individual shape complexity using only fast feed-forward inference. Our method operates on general latent representations and improves rate--distortion for both meshes and signed distance fields.
Code is available at \href{https://github.com/voilalab/varlen-3D}{github.com/voilalab/varlen-3D}.

\end{abstract}

\begin{keywords}
variable-length compression, tokenization, 3D representation, mesh, signed distance function
\end{keywords}

\vspace{-3mm}
\section{Introduction}
\label{sec:intro}
Representing 3D shapes is an essential step in many applications, including 3D reconstruction, 3D generation, physical simulation, gaming, augmented reality, and manufacturing design.
The shapes in these applications differ widely in complexity.
Most current 3D representations, such as voxels and meshes, represent shapes at full fidelity and therefore require substantial storage regardless of shape complexity.
Storing, transmitting, and processing these representations without any compression can thus be costly.

Traditional 3D compression methods~\cite{edgebreaker, pcc, vpcc, draco} exploit specific knowledge about the shape or the representation to achieve compression.
As a result, they cannot readily adapt to new shape representations out-of-the-box or leverage training examples to improve compression.
Recent work has explored learning-based methods in two directions.
Per-shape methods fit a small model to each shape independently~\cite{nglod,bacon,ngf,qndf,hns}.
These methods do not require any training examples of other shapes or share any parameters between different shapes.
However, they cannot exploit prior knowledge about what shapes typically look like, which limits how much they can compress.
Moreover, they require computationally costly per-shape optimization at inference time.
An alternative is to train an encoder and a decoder on a large collection of shapes~\cite{meshgpt,10.1145/3592442,cofie}, so that the encoder outputs a compact representation while the decoded shapes maintain high fidelity.
Such learned compressors hold the promise of improving with more data and compute, but their compression ratio is usually fixed as a hyperparameter before training.
Every shape is therefore compressed at the same ratio, regardless of its complexity, which limits the achievable rate--distortion tradeoff across shapes of variable complexity.

We propose a learning-based variable-length compressor that retrofits any pretrained shape autoencoder.
Specifically, we use a frozen encoder to map a shape to a latent vector, and train a small hyper-autoencoder that maps this latent vector into an ordered sequence of tokens.
A regression decoder then maps any prefix of this token sequence to approximate the full latent vector, and then the full shape, with quality improving as more tokens are kept.
At inference time, users can choose how many tokens to keep per shape, guided by quality metrics computed in either the shape or latent space so more bits can be allocated to more complex shapes.
Since the underlying autoencoder is frozen, our method can be combined with any existing shape autoencoder
without retraining it.

We demonstrate that our method applies to different 3D representations, such as meshes and signed distance functions (SDFs), and that it can be supervised in both the latent space and the shape space.
We evaluate on ShapeNet~\cite{shapenet} utilizing both supervision strategies and compare rate-distortion, measured as Chamfer distance versus compression ratio, against per-shape and dataset-level compression baselines.
We find that our method achieves a better rate-distortion tradeoff than both families of baselines (in the absence of quantization), with the property that the number of tokens needed to reach a target quality increases with shape complexity.
Ablation studies show that our regression decoder outperforms a FlexTok-style flow decoder \cite{flextok}, with most of the gain coming from training on the downstream shape losses rather than intermediate latent losses.

In summary, our contributions are:
\begin{itemize}[nosep]
    \item A novel 3D compression framework that reuses pretrained shape autoencoders and allows the compression ratio to be chosen per shape at inference time without retraining, and
    \item State-of-the-art rate--distortion on meshes and SDFs, compared to both per-shape and dataset-level unquantized compression baselines.
\end{itemize}

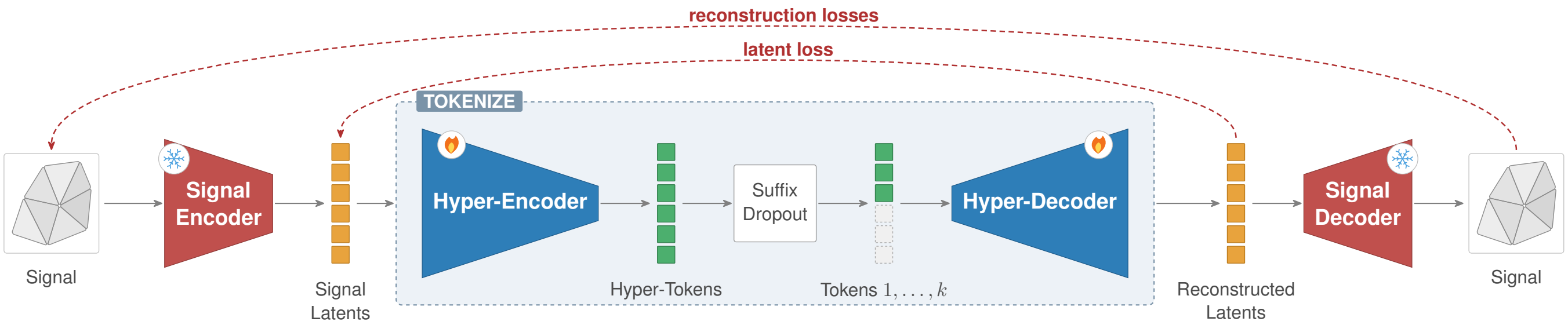
\begin{figure*}[h]
  \centering
  \resizebox{\textwidth}{!}{\providecommand{\figfontA}{\sffamily\fontsize{13}{15}\selectfont}%
\providecommand{\figfontB}{\sffamily\fontsize{11}{13}\selectfont}%
\definecolor{encblue}  {HTML}{2E7EB8}%
\definecolor{encbluedk}{HTML}{1F5C8B}%
\definecolor{gcnred}   {HTML}{C0504D}%
\definecolor{gcnreddk} {HTML}{8F3A38}%
\definecolor{tokorange}{HTML}{E8A33D}%
\definecolor{tokordk}  {HTML}{B87719}%
\definecolor{tokgreen} {HTML}{4CAF6E}%
\definecolor{tokgrndk} {HTML}{2E7A48}%
\definecolor{lossred}  {HTML}{B03030}%
\definecolor{grouppale}{HTML}{EDF2F7}%
\definecolor{groupline}{HTML}{7A93A8}%
\definecolor{inkgray}  {HTML}{4A4A4A}%
\providecommand{\meshglyph}{}\renewcommand{\meshglyph}{%
\tikz[baseline=(current bounding box.center), scale=1.55, line width=0.45pt,%
      draw=black!45]{%
  \fill[black!7]  (0,0.15)--(0.45,0)--(0.15,0.6)--cycle;%
  \fill[black!13] (0.45,0)--(0.6,0.5)--(0.15,0.6)--cycle;%
  \fill[black!10] (0.45,0)--(0.95,0.2)--(0.6,0.5)--cycle;%
  \fill[black!16] (0.95,0.2)--(1.0,0.72)--(0.6,0.5)--cycle;%
  \fill[black!11] (0.15,0.6)--(0.6,0.5)--(0.4,1.0)--cycle;%
  \fill[black!7]  (0.6,0.5)--(0.86,1.06)--(0.4,1.0)--cycle;%
  \fill[black!14] (0.6,0.5)--(1.0,0.72)--(0.86,1.06)--cycle;%
  \draw (0,0.15)--(0.45,0)--(0.15,0.6)--cycle;%
  \draw (0.45,0)--(0.6,0.5)--(0.15,0.6)--cycle;%
  \draw (0.45,0)--(0.95,0.2)--(0.6,0.5)--cycle;%
  \draw (0.95,0.2)--(1.0,0.72)--(0.6,0.5)--cycle;%
  \draw (0.15,0.6)--(0.6,0.5)--(0.4,1.0)--cycle;%
  \draw (0.6,0.5)--(0.86,1.06)--(0.4,1.0)--cycle;%
  \draw (0.6,0.5)--(1.0,0.72)--(0.86,1.06)--cycle;%
}}%
\providecommand{\stackglyph}[2]{}\renewcommand{\stackglyph}[2]{%
\tikz[baseline=(current bounding box.center), line width=0.4pt]{%
  \foreach \i in {1,...,#2}{%
    \ifnum\i>#1%
      \draw[draw=black!35, fill=black!6, dash pattern=on 1.2pt off 1.2pt,%
            rounded corners=0.3pt]%
            (0,-0.40*\i) rectangle ++(0.34,0.33);%
    \else%
      \draw[draw=tokordk, fill=tokorange, rounded corners=0.3pt]%
            (0,-0.40*\i) rectangle ++(0.34,0.33);%
    \fi}%
}}%
\providecommand{\stackglyphg}[2]{}\renewcommand{\stackglyphg}[2]{%
\tikz[baseline=(current bounding box.center), line width=0.4pt]{%
  \foreach \i in {1,...,#2}{%
    \ifnum\i>#1%
      \draw[draw=black!35, fill=black!6, dash pattern=on 1.2pt off 1.2pt,%
            rounded corners=0.3pt]%
            (0,-0.40*\i) rectangle ++(0.34,0.33);%
    \else%
      \draw[draw=tokgrndk, fill=tokgreen, rounded corners=0.3pt]%
            (0,-0.40*\i) rectangle ++(0.34,0.33);%
    \fi}%
}}%
\definecolor{iceblue}    {HTML}{4FA3E0}%
\definecolor{flameorange}{HTML}{F2711C}%
\definecolor{flameyellow}{HTML}{FFD54F}%
\providecommand{\snowglyph}{}\renewcommand{\snowglyph}{%
\tikz[baseline=(current bounding box.center), line width=0.55pt, draw=iceblue, line cap=round]{%
  \foreach \a in {0,60,120}{\draw (\a:-1.5mm) -- (\a:1.5mm);}%
  \foreach \a in {0,60,...,300}{%
    \draw (\a:1.5mm) -- ++(\a+40:0.55mm); \draw (\a:1.5mm) -- ++(\a-40:0.55mm);%
    \draw (\a:0.8mm) -- ++(\a+50:0.4mm);  \draw (\a:0.8mm) -- ++(\a-50:0.4mm);}%
}}%
\providecommand{\flameglyph}{}\renewcommand{\flameglyph}{%
\tikz[baseline=(current bounding box.center)]{%
  \fill[flameorange] (0,-1.6mm) .. controls (-1.7mm,-0.6mm) and (-1.5mm,0.9mm) .. (-0.3mm,1.7mm)%
     .. controls (0.1mm,1.0mm) and (0.6mm,1.2mm) .. (0.9mm,1.9mm)%
     .. controls (1.9mm,0.9mm) and (1.6mm,-0.6mm) .. (0,-1.6mm) -- cycle;%
  \fill[flameyellow] (0,-1.5mm) .. controls (-0.8mm,-0.6mm) and (-0.6mm,0.2mm) .. (0.1mm,0.6mm)%
     .. controls (0.8mm,0.2mm) and (0.8mm,-0.6mm) .. (0,-1.5mm) -- cycle;%
}}%
\begin{tikzpicture}[
  font=\figfontA,
  node distance=12mm,
  block/.style={rounded corners=1.2pt, align=center, text=white,
                font=\figfontA\bfseries, inner sep=3pt,
                minimum height=15mm, minimum width=17mm},
  enc/.style={trapezium, trapezium angle=72, shape border rotate=270,
              align=center, text=white, font=\figfontA\bfseries,
              inner xsep=6pt, inner ysep=4pt, minimum height=21mm},
  dec/.style={trapezium, trapezium angle=72, shape border rotate=90,
              align=center, text=white, font=\figfontA\bfseries,
              inner xsep=6pt, inner ysep=4pt, minimum height=21mm},
  op/.style={rounded corners=1.2pt, draw=inkgray!60, fill=white, align=center,
             font=\figfontB, text=inkgray, inner sep=4pt,
             minimum height=15mm, minimum width=16mm},
  glyph/.style={inner sep=1pt, outer sep=2pt},
  lbl/.style={font=\figfontB, text=inkgray, align=center, inner sep=0pt},
  card/.style={rounded corners=1.5pt, draw=black!25, fill=white, inner sep=4pt},
  arr/.style={-{Stealth[length=2.2mm,width=1.7mm,round]}, line width=0.7pt,
              draw=inkgray!70, shorten >=1pt, shorten <=1pt}
,
  badge/.style={circle, draw=black!25, fill=white, inner sep=0.4pt,
                minimum size=4.6mm, line width=0.4pt}
]

\node[glyph, card] (A) {\meshglyph};
\node[lbl, below=2.5mm of A] {Signal};

\node[enc, fill=gcnred, draw=gcnreddk, right=of A] (B)
  {Signal\\Encoder};

\node[glyph, right=10.3mm of B] (C) {\stackglyph{6}{6}};
\node[lbl, below=2.5mm of C] {Signal\\Latents};

\node[enc, fill=encblue, draw=encbluedk, right=13mm of C] (D) {Hyper-Encoder};

\node[glyph, right=10.3mm of D] (E) {\stackglyphg{6}{6}};
\node[lbl, below=2.5mm of E] {Hyper-Tokens};

\node[op, right=10.3mm of E] (F) {Suffix\\Dropout};

\node[glyph, right=10.3mm of F] (G) {\stackglyphg{3}{6}};
\node[lbl, below=2.5mm of G] {Tokens $1,\dots,k$};

\node[dec, fill=encblue, draw=encbluedk, right=10.3mm of G] (H) {Hyper-Decoder};

\begin{scope}[on background layer]
  \node[fit=(D)(E)(F)(G)(H), inner xsep=5mm, inner ysep=7mm,
        rounded corners=3pt, draw=groupline, dash pattern=on 2pt off 2pt,
        line width=0.7pt, fill=grouppale] (AE) {};
\end{scope}
\node[font=\figfontB\bfseries, text=groupline!80!black,
      anchor=west, xshift=3mm] at (AE.north west|-AE.north) {};
\node[anchor=west, font=\figfontB\bfseries, text=white, fill=groupline,
      rounded corners=1.2pt, inner xsep=4pt, inner ysep=2pt]
      at ($(AE.north west)+(4mm,0)$) {TOKENIZE};

\node[glyph, right=13mm of AE] (I) {\stackglyph{6}{6}};
\node[lbl, below=2.5mm of I] {Reconstructed\\Latents};

\node[dec, fill=gcnred, draw=gcnreddk, right=10.3mm of I] (K) {Signal\\Decoder};%

\node[glyph, card, right=10.3mm of K] (L) {\meshglyph};
\node[lbl, below=2.5mm of L] {Signal};

\node[badge] at ($(B.north west)+(1.9mm,-1.9mm)$) {\snowglyph};
\node[badge] at ($(K.north east)+(-1.9mm,-1.9mm)$) {\snowglyph};
\node[badge] at ($(D.north west)+(1.9mm,-1.9mm)$) {\flameglyph};
\node[badge] at ($(H.north east)+(-1.9mm,-1.9mm)$) {\flameglyph};

\draw[arr] (A) -- (B);
\draw[arr] (B) -- (C);
\draw[arr] (C) -- (AE.west);
\draw[arr] (D) -- (E);
\draw[arr] (E) -- (F);
\draw[arr] (F) -- (G);
\draw[arr] (G) -- (H);
\draw[arr] (AE.east) -- (I);
\draw[arr] (I) -- (K);
\draw[arr] (K) -- (L);

\draw[-{Stealth[length=2.2mm,width=1.7mm,round]}, line width=0.8pt,
      draw=lossred, dash pattern=on 3pt off 2pt]
  (I.north) .. controls ($(I.north)+(0,2.0)$) and ($(C.north)+(0,2.0)$) ..
  node[pos=0.5, above=-1pt, font=\figfontB\bfseries, text=lossred]
  {latent loss} (C.north);
\draw[-{Stealth[length=2.2mm,width=1.7mm,round]}, line width=0.8pt,
      draw=lossred, dash pattern=on 3pt off 2pt]
  (L.north) .. controls ($(L.north)+(0,3.2)$) and ($(A.north)+(0,3.2)$) ..
  node[pos=0.5, above=-1pt, font=\figfontB\bfseries, text=lossred]
  {reconstruction losses} (A.north);

\end{tikzpicture}%
}
  \caption{Overview of our variable-length hyper-autoencoder, shown for an arbitrary 3D signal. A pretrained signal
  encoder
  maps the input signal to
  per-element latents; our suffix-dropout autoencoder compresses them into an
  ordered sequence of tokens (green), where any prefix of $k$ tokens suffices to
  reconstruct the latents, which are
  decoded back to the
  signal space. Dashed arrows show the two supervision paths: the latent loss
  (token $\to$ token) and the downstream reconstruction losses
  (signal $\to$ signal). Note that backpropagation along the signal-level path is
  possible for our regression hyper-decoder but not for a flow-based hyper-decoder, as is used in FlexTok \cite{flextok}.}
  \label{fig:meshgpt_tokenizer}
\end{figure*}

\section{Related Works}
\label{sec:relatedworks}

\noindent\textbf{3D Compression.}
Traditional compression techniques for 3D assets are highly specialized to the type of representation, such as point clouds \cite{pcc} and meshes \cite{vpcc}, and are designed for compatibility with existing 2D image and video codecs rather than for optimal compression performance.
Recently,
per-shape optimization frameworks have been proposed for compressing 3D geometries, often signed distance functions (SDFs) \cite{nglod, cofie} or displacement fields \cite{ngf, qndf, hns}, as implicit neural representations.
While these methods often yield better rate--distortion tradeoffs than traditional codecs, they take much longer to encode each shape or change the encoded bit-rate for a given shape, and
cannot exploit cross-shape or semantic redundancies.

\noindent\textbf{Variable-Length Tokenization.}
Recent work has explored variable-length or hierarchical image and video representations for applications such as autoregressive generation and retrieval. Matryoshka Representation Learning
\cite{kusupati2024matryoshkarepresentationlearning} introduced variable-length latent image representations, while TiTok \cite{titok} adapted this idea to tokenization by using suffix token dropout to learn ordered token sequences whose prefixes provide progressively higher-fidelity reconstructions. ElasticTok \cite{elastictok} extended this approach to video, training a variable-length VQ-VAE from scratch. Both TiTok and ElasticTok demonstrate autoregressive generation from ordered tokens, but require costly training of specialized autoencoders on large datasets.

FlexTok \cite{flextok} proposed an alternative formulation for variable length tokenization, leveraging a pretrained variational autoencoder (VAE). FlexTok converts the fixed-size latents of the pretrained VAE into an ordered sequence of tokens using suffix dropout,
so that any variable-length token prefix can be used as context-tokens for a rectified-flow-based decoder \cite{flux} that is trained to approximate the tokens of the original VAE.
While this formulation is suitable for generating feasible shapes even when few context tokens are available, it falls short when stricter reconstruction quality criteria are imposed.
Additionally,
FlexTok requires expensive retraining of their generative Flux VAE decoder \cite{flux}.

Our proposed framework simultaneously addresses the limitations of these families of variable-length 3D compression methods by re-purposing pretrained shape autoencoders and tokenizers for the task of variable bit-rate compression. Existing shape encoders such as Mesh-GPT \cite{meshgpt}, 3DShape2VecSet \cite{10.1145/3592442}, LaGeM \cite{lagem}, etc. are trained on vast 3D-asset datasets such as ShapeNet \cite{shapenet} and Objaverse-XL \cite{objaverse}, and consequently have learned rich geometric priors which our compression pipeline can seamlessly inherit.
Our framework operates in the latent/token space of these foundational autoencoder models, which enables it to leverage pretrained models without retraining them and makes it modality agnostic (\emph{e.g.}, applicable to SDFs, meshes, and images) as long as there is a pretrained autoencoder for each modality.
Our method can seamlessly adapt to additional modalities as corresponding foundation autoencoders become available.%

\begin{figure*}[t]
  \centering
  \providecommand{\stile}[1]{\includegraphics[width=0.093\textwidth]{figures/tiles/#1.png}}%
\providecommand{\shead}[1]{{\scriptsize\textbf{#1}}}%
\providecommand{\srowlab}[1]{\rotatebox{90}{\makebox[0.093\textwidth][c]{\scriptsize\textbf{#1}}}}%
\providecommand{\stileim}[1]{\includegraphics[width=0.132\textwidth]{figures/tiles/#1.png}}%
\providecommand{\srowlabim}[1]{\rotatebox{90}{\makebox[0.132\textwidth][c]{\scriptsize\textbf{#1}}}}%
\setlength{\tabcolsep}{1pt}%
\centering

\newlength{\dtile}\setlength{\dtile}{\dimexpr(\textwidth-44pt)/8\relax}%
\providecommand{\stiled}[1]{\includegraphics[width=\dtile]{figures/tiles/#1.png}}%
\providecommand{\srowlabd}[1]{\rotatebox{90}{\makebox[\dtile][c]{\scriptsize\textbf{#1}}}}%
\begin{tabular}{@{}ccccc@{\hspace{8pt}}ccccc@{}}
  & \shead{\textmd{Ground Truth}} & \shead{$k{=}32$} & \shead{$k{=}64$} & \shead{$k{=}256$}
 & & \shead{\textmd{Ground Truth}} & \shead{$k{=}64$} & \shead{$k{=}128$} & \shead{$k{=}256$} \\
\srowlabd{SDF} & \stiled{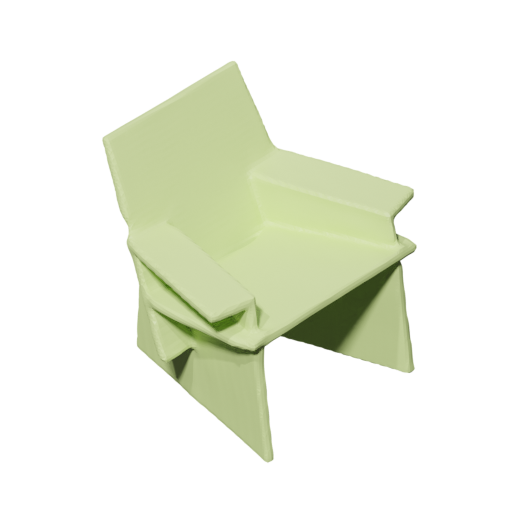} & \stiled{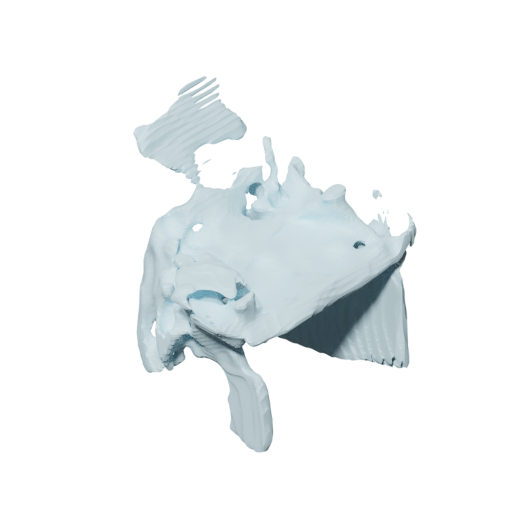} & \stiled{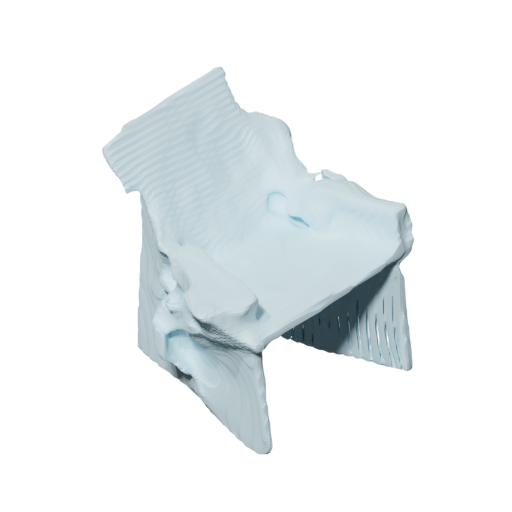} & \stiled{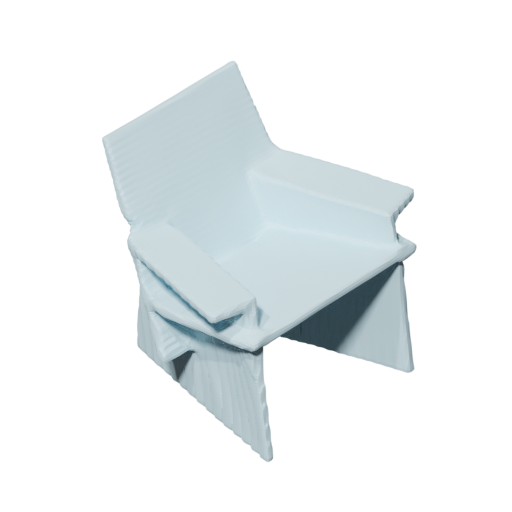}
  & \srowlabd{Mesh} & \stiled{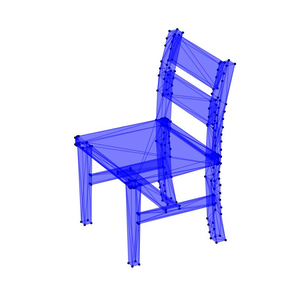} & \stiled{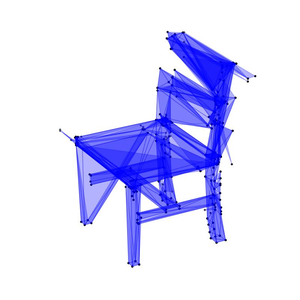} & \stiled{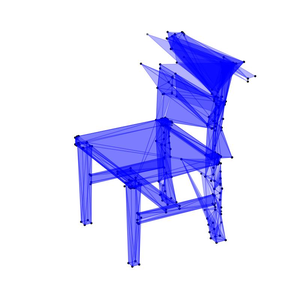} & \stiled{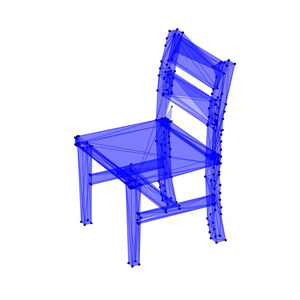} \\
\end{tabular}
\\[5pt]
  \includegraphics[width=0.24\textwidth]{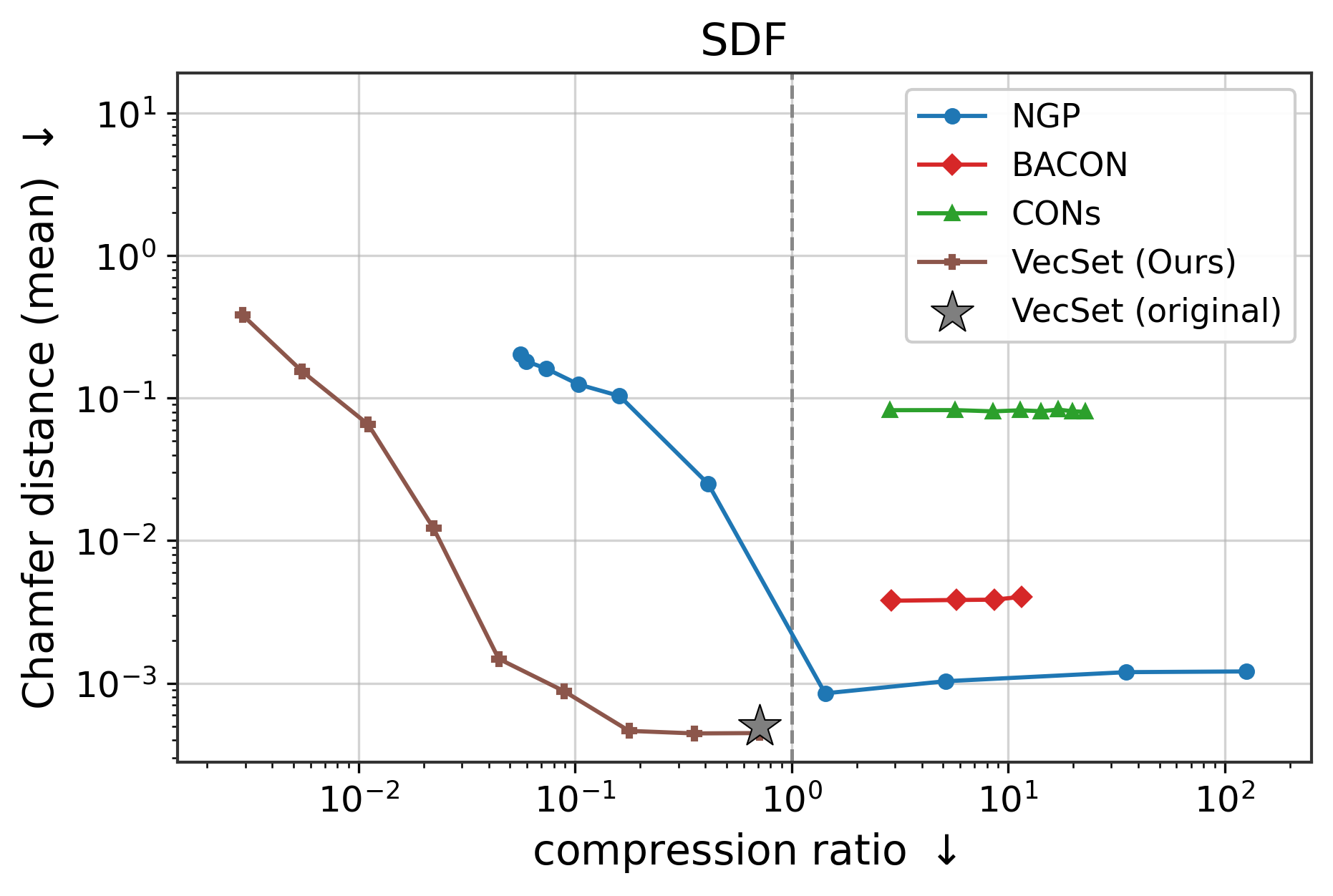}\hfill
  \includegraphics[width=0.24\textwidth]{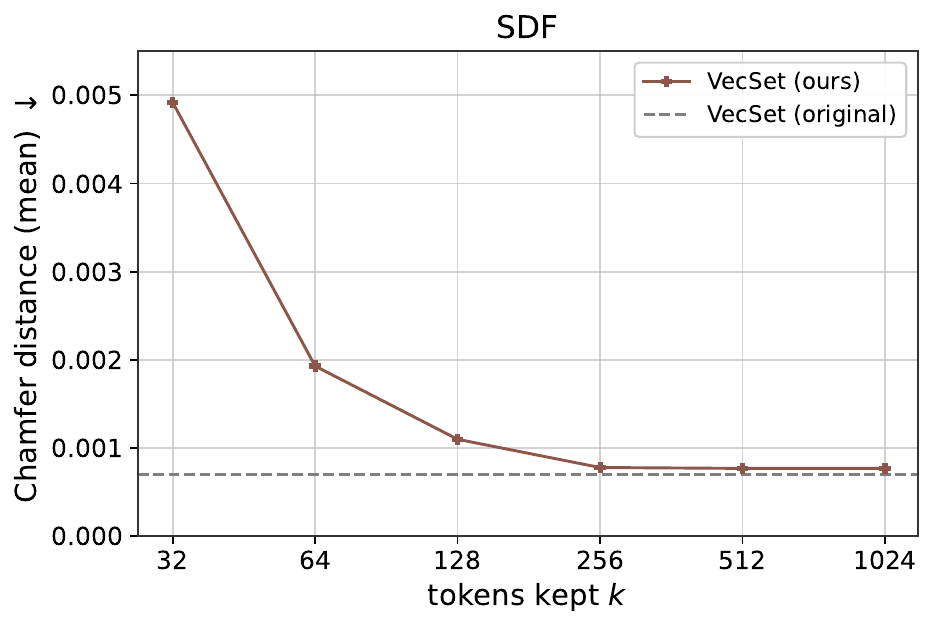}\hfill
  \includegraphics[width=0.24\textwidth]{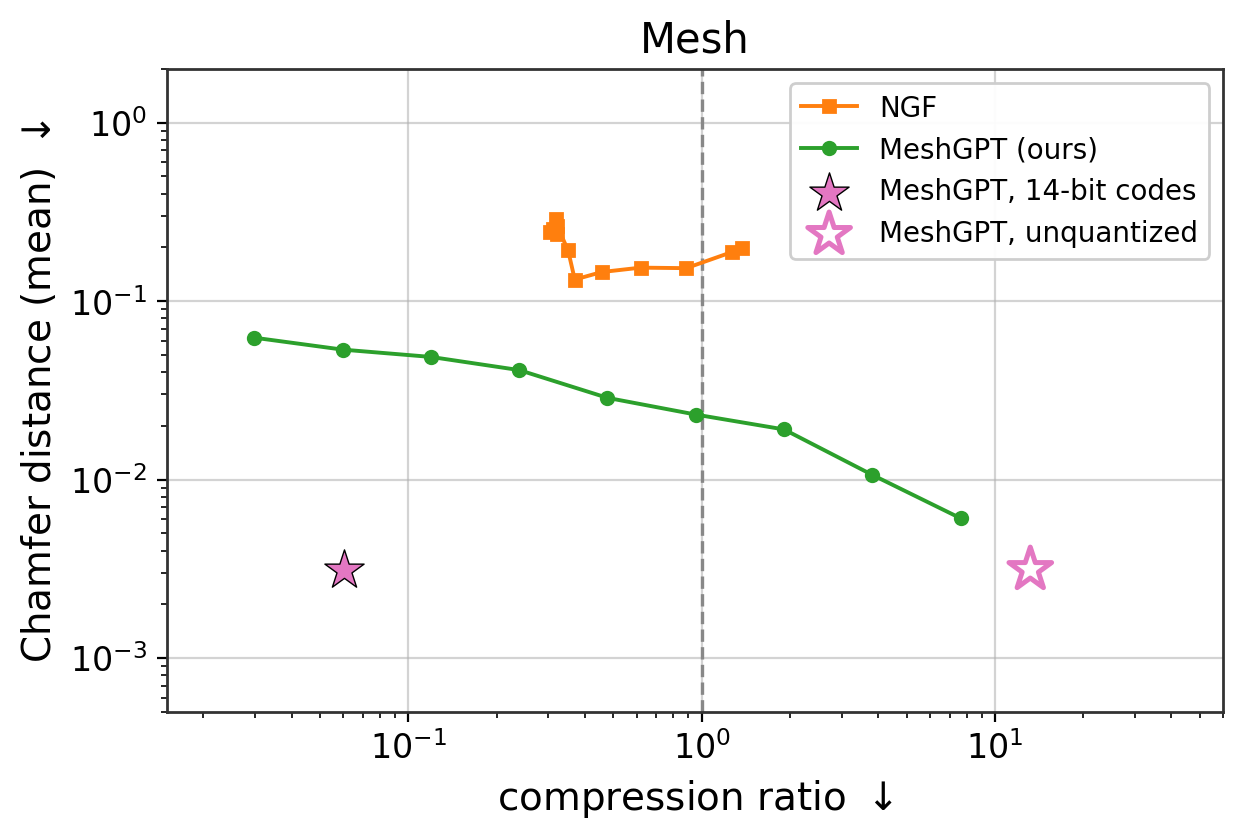}\hfill
  \includegraphics[width=0.24\textwidth]{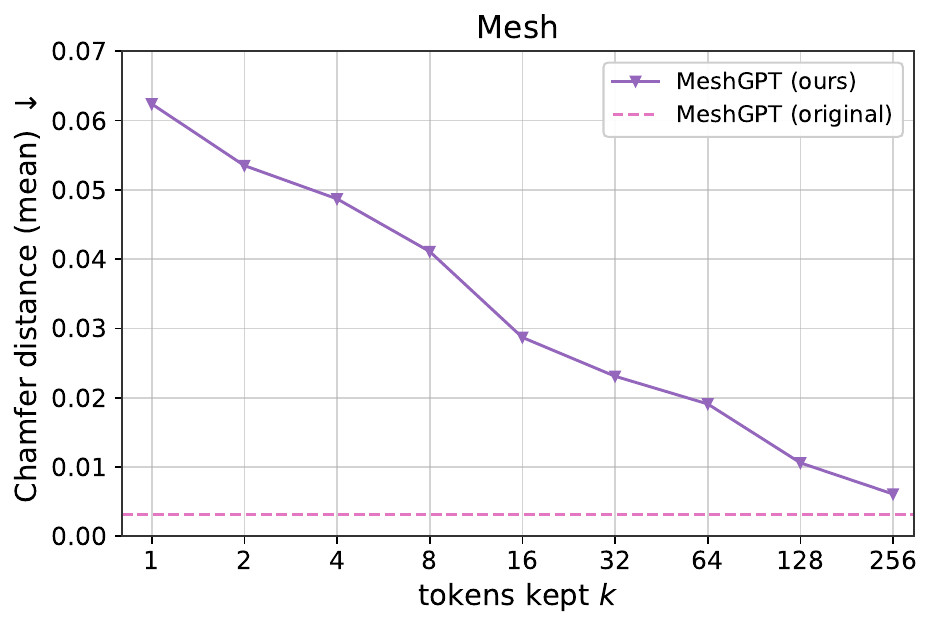}\hfill
  \caption{Variable-length compression of
  SDF and MeshGPT latents.
  Top: Sample SDF and mesh reconstructions, each decoded from the first $k$ tokens. As $k$ increases, so does the fidelity compared to the ground truth.
  Bottom, column 1: Baseline comparison on rate-distortion curve for SDF/implicit methods. Our method represents a Pareto frontier.
  Bottom, column 2: Chamfer distance vs.\ $k$ for SDF compression.
  Bottom, column 3: Baseline comparison of rate-distortion curve for mesh compression. We show for comparison compression ratios for both quantized and unquantized versions of MeshGPT---our method is unquantized, so our curve is shifted to the right.
  Bottom, column 4: Chamfer distance vs.\ $k$ for mesh compression.}
  \label{fig:results_main}
\end{figure*}

\vspace{-2mm}
\section{Method}
\label{sec:method}

Our method assumes access to a pretrained signal encoder
$\mathcal{E}: \mathcal{S} \rightarrow \mathcal{Z}$ that encodes a signal $s\in\mathcal{S}$ of a specific type of modality (\emph{e.g.}, $\mathcal{S}$ could be the space of meshes, SDFs, or images) into a latent vector or token set $z \in \mathcal{Z}$, and its corresponding decoder $\mathcal{D}: \mathcal{Z}\rightarrow\mathcal{S}$ that reconstructs the latent code back into an approximation of the original signal.
Our goal is to convert $z$ to an \emph{ordered} set of tokens $\{t_1, t_2, \dots, t_K\} \in \mathbb{T}$, where $t_i \in {\mathbb{R}^d}$, so that any prefix of tokens $\{t_1,\dots, t_k\}$ can be decoded into an approximation of the original latent vector $z$, with $k\leq K$ and fidelity increasing with $k$.

We train a \textit{hyper-autoencoder} to perform this transformation, as shown in \Cref{fig:meshgpt_tokenizer}.
Our input $z$ is a flat list of tokens in the form $\{z_1, z_2, \dots, z_Z\}$.
We apply Perceiver-style cross-attention \cite{perceiver}, where the keys and values are derived from $z$ as shown in \Cref{eq:encoder_attn} and the queries come from a learned vector set $Q^{(e)}:=\{q_1^{(e)}, q_2^{(e)}, \dots, q_K^{(e)}\}$ that is fixed once the hyper-encoder is trained.
The outputs of this cross attention (CA) operation, which we call hyper-tokens, form a sequence $\{t_1, t_2, \dots, t_K\}$ of the same length $K$ as the learned query set.
Compactly, this hyper-encoding operation is:
\begin{equation}\label{eq:encoder_attn}
    t_i = \text{CA}(\text{query}=Q^{(e)}, \text{key}=z_{1:Z}) \quad\forall i \in \{1,...,K\},
\end{equation}
Next, these produced hyper-tokens $t_i$ undergo a sequence of causal self-attention layers.
The ordering of the hyper-tokens results from both the inherent ordering in the learned queries and the causal nature of the self-attention we apply to the hyper-tokens, which encourages earlier tokens to capture as much structure as possible.

Finally, our hyper-decoder reconstructs an approximation of the full original token/latent set even from a truncated prefix $\{t_1,\dots,t_k\}$ of the hyper-token set.
We perform this decoding using Perceiver-style cross attention, which uses the hyper-token sequence to derive the keys and values and uses a separate set of $Z$ learned decoder tokens $Q^{(d)}:=\{q_1^{(d)}, q_2^{(d)}, \dots, q_Z^{(d)}\}$ as queries.
The reconstructed token set $\{\Tilde{z}_1, \Tilde{z}_2, \dots, \Tilde{z}_Z\}$ always has the same length $Z$ irrespective of the number of suffix hyper-tokens that are dropped, due to the fixed length of the learned query tokens $Q^{(d)}$.
Notationally,
\begin{equation}
    \hspace{-2mm}\Tilde{z}_i^{(k)} = \text{CA}(\text{query}=Q^{(d)}, \text{key}=t_{1:k}) \quad\forall i \in \{1,..., Z\},
\end{equation}
where $k\leq K$ is the length of the truncated hyper-token sequence.
The superscript $(k)$ in $\Tilde{z}_i^{(k)}$ denotes the number of hyper-tokens used to reconstruct an approximation of the original latent set element $z_i$.
We denote the hyper-autoencoder reconstruction of the full latents using $k$ hyper-tokens as $\Tilde{z}^{(k)} = \text{HAE}_k(z)$.

We train the hyper-autoencoder in two self-supervised stages: latent-space supervision and signal-space finetuning.
In the first stage, given a dataset of signals in $\mathcal{S}$, we minimize the latent-space loss:
\begin{equation}
    \mathcal{L}_\text{latent} = \mathbb{E}_{s\sim \mathcal{S}}\mathbb{E}_{k\leq K}\left[\|\mathcal{E}(s) - \text{HAE}_k(\mathcal{E}(s))\|_2^2\right].
\end{equation}
In the second stage, we finetune the hyper-autoencoder with signal-space losses
to retain the performance of the pretrained signal autoencoder:
\begin{equation}
    \mathcal{L}_\text{signal} = \mathbb{E}_{s\sim \mathcal{S}}\mathbb{E}_{k\leq K}\text{SignalLoss}(s,  \mathcal{D}(\text{HAE}_k(\mathcal{E}(s)))].
\end{equation}
Here, SignalLoss could be any loss function that is specific to autoencoder training for the given modality, for example mean squared error over SDF samples.

Including this initial stage of latent-space training offers two benefits:
(i) latent-space supervision is faster, as it avoids the expensive gradient flow through the signal decoder $\mathcal{D}(\cdot)$; (ii) latent-space supervision is more stable at early stages of training, when the initial latents are far from the pretrained decoder's training distribution and thus signal-space error can produce uninformative gradients.

Because these loss functions encourage the hyper-decoder to be able to approximate the full original latent and signal even from an arbitrarily sized prefix sequence of hyper-tokens, the hyper-encoder naturally learns to encode coarse structure of the signals in hyper-tokens of earlier indices while reserving the higher-index hyper-tokens to enrich the reconstructions with higher frequency or fine-grained details.
Once the hyper-autoencoder is trained, a given signal can be converted to a sequence of hyper-tokens in a feed-forward manner which can either be truncated arbitrarily after encoding, or have its suffix dropped out to obtain compressed representations with varying bit-rates.

\section{Results}
\label{sec:results}

We experimentally evaluate our method, demonstrating that given arbitrary latent representations of 3D shapes, it induces a coarse-to-fine variable-length compression.
We compare our method against relevant baselines and show that it achieves state-of-the-art compression for 3D representations, including both meshes and SDFs by tokenizing pretrained latent spaces of MeshGPT \cite{meshgpt} and 3DShape2VecSet \cite{10.1145/3592442} models respectively.

In Figure~\ref{fig:results_main}, we show sample 3D reconstructions of both meshes and SDFs while varying $k$, the number of hyper-tokens retained. We visually observe that as $k$ increases, so does the amount of information preserved compared to the ground truth, with the highest $k$ resulting in the most faithful reconstruction.
This relationship is also visible in the second and fourth plots in the bottom row of Figure~\ref{fig:results_main}, where we compare the reconstruction quality (Chamfer distance) vs. the number of hyper-tokens $k$ for both modalities. In each case, we observe an inverse relationship.

In the first and third plots of Figure~\ref{fig:results_main}, we evaluate rate--distortion for our method as well as modality-specific baselines for both meshes and SDFs.
For each modality, we graph our method's reconstruction error (Chamfer distance) vs. compression ratio, alongside other variable-length compression baselines and the fixed-length signal autoencoder (VecSet for SDF and MeshGPT for mesh) from which our hyper-autoencoder is derived.
We observe that for both modalities our method lies on the Pareto frontier in fidelity vs. compression ratio, and that the original signal autoencoder does not exceed our method's performance, showing that our hyper-autoencoder does not induce performance degradation.

\begin{figure}
    \includegraphics[width=0.48\textwidth]{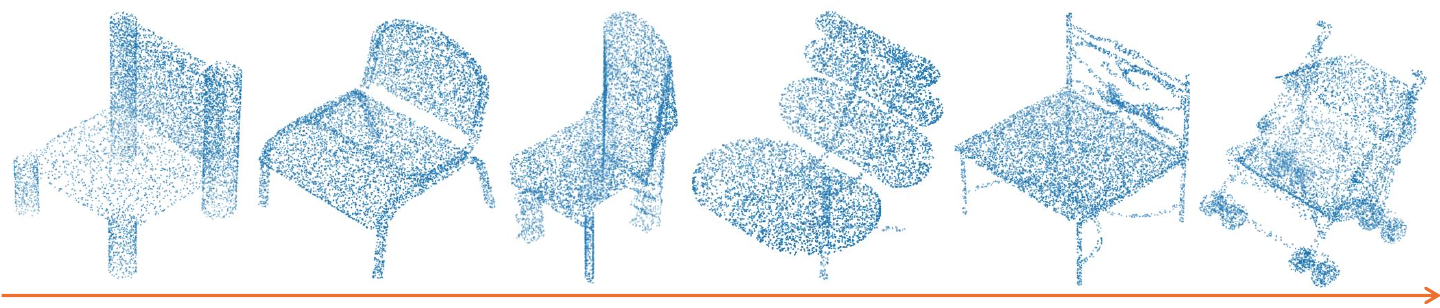}
    \caption{Number of hyper-tokens needed to achieve a target MSE threshold in latent space increases with the ``geometric complexity'' of shapes.}
    \label{fig:complexityvstokens}
\end{figure}

In \Cref{fig:complexityvstokens}, we visualize example shapes from our SDF experiments stratified by the number of hyper-tokens needed to achieve a threshold error in reconstructing the original shape latents. We observe that shapes that require more tokens (larger $k$) are visually more complex, indicating that our hyper-autoencoder learns to compress simpler shapes using fewer tokens, achieving variable-length compression.

\subsection{Ablations}

We perform ablation studies on key elements of our method. In particular, compared to FlexTok \cite{flextok}, another variable-length tokenizer, we use a regression hyper-decoder whereas FlexTok uses a flow-based generative decoder. For the compression task, we find that this is a key difference because it enables our method to train on both signal- and latent-space losses, rather than only latent-space losses.
Our method can train on downstream signal-space losses through its regression decoder, but not through a flow decoder, as we cannot backpropagate through ODE sampling, whereas regression can be directly supervised by geometric losses. We hypothesize that the significant increase in performance when training on these signal-space losses is due to the realtive lack of strong, perceptually aligned latent representations in the 3D domain vs. the image domain for which FlexTok was designed.

To isolate the strength of these effects, we compare three decoders: (1) our regression decoder trained with the downstream reconstruction losses, (2) the same regression decoder trained with the latent loss only, and (3) a FlexTok-style flow decoder.
We show the results in \Cref{fig:ablation_curves}, and observe that the inclusion of signal-space reconstruction loss provides substantial improvements in fidelity.

\newcommand{\plotcrop}[1]{%
\adjincludegraphics[trim={{0.0\width} {0.0\height} {0.0\width} {0.0\height}}, clip, width=0.42\textwidth]{#1}%
}
\begin{figure}[t]
  \centering
  \plotcrop{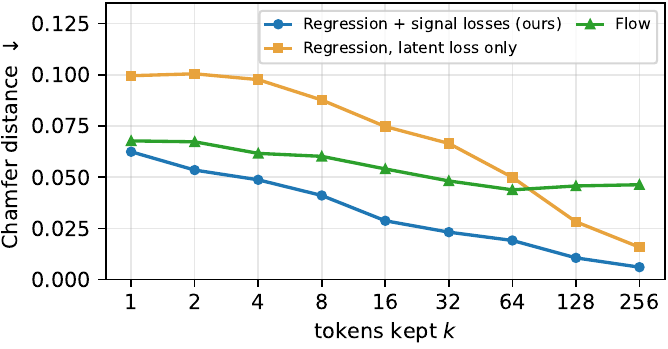}\hfill
  \caption{Decoder ablation. Chamfer distance vs.\ $k$ on meshes for our regression decoder, the same regression decoder trained with the latent loss only,
  and the flow decoder.
  At low $k$ the flow
  decoder returns a sharper but wrong signal; at high $k$ the regression decoders are more faithful.
  }
  \label{fig:ablation_curves}
\end{figure}

\section{Conclusions}
\label{sec:conclusions}
We propose a data-driven variable-length compression method for 3D representations.
Our method can operate on the latent-space of any pretrained signal autoencoder and allocate fewer hyper-tokens to simpler signals and more to complex signals. Our approach demonstrates state-of-the-art compression results for both mesh and SDF representations of 3D shapes.
Interesting future directions include extending our method to other signal modalities and 3D representations.

\vfill\pagebreak

\bibliographystyle{IEEEbib}
\bibliography{refs}

\end{document}